\documentclass[letterpaper]{article} 
\usepackage{aaai20}  
\usepackage{times}  
\usepackage{helvet} 
\usepackage{courier}  
\usepackage[hyphens]{url}  
\usepackage{graphicx} 
\usepackage{graphicx}  
\usepackage{times}
\usepackage{epsfig}
\usepackage{graphicx}
\usepackage{amsmath}
\usepackage{amssymb}
\usepackage{array}
\usepackage{booktabs} 
\usepackage{multirow} 
\usepackage{multicol} 
\usepackage{arydshln}
\usepackage[pagebackref=true,breaklinks=true,letterpaper=true,colorlinks,bookmarks=false]{hyperref}
\usepackage{bm}

\title{STA: Adversarial Attacks on Siamese Trackers}
\author{Xugang Wu\textsuperscript{\rm 1}, XiaoPing Wang\textsuperscript{\rm 2}, XuZhou\textsuperscript{\rm 1}, Songlei Jian\textsuperscript{\rm 1}, Kai Lu\textsuperscript{\rm 1}\\
\textsuperscript{\rm 1}National University of Defense Technology\\ 
\textsuperscript{\rm 2}Hunan Univeristy\\
xugangwu95@gmail.com
}

\begin{document}

\maketitle

\begin{abstract}
Recently, the majority of visual trackers adopt Convolutional Neural Network (CNN) as their backbone to achieve high tracking accuracy. However, less attention has been paid to the potential adversarial threats brought by CNN, including Siamese network. 
   
In this paper, we first analyze the existing vulnerabilities in Siamese trackers and propose the requirements for a successful adversarial attack. On this basis, we formulate the adversarial generation problem and propose an end-to-end pipeline to generate a perturbed texture map for the 3D object that causes the trackers to fail. Finally, we conduct thorough experiments to verify the effectiveness of our algorithm. Experiment results show that adversarial examples generated by our algorithm can successfully lower the tracking accuracy of victim trackers and even make them drift off. To the best of our knowledge, this is the first work to generate 3D adversarial examples on visual trackers.

\end{abstract}

\section{Introduction}
Visual object tracking, which aims at locating the trajectory of a target in an image sequence given its initial state, is one of the fundamental problems in computer vision. In recent years, visual trackers have achieved great performance improvement due to the application of Convolutional Neural Network (CNN), a powerful tool for learning representations of images\cite{krizhevsky2012imagenet}. Among these CNN based trackers, Siamese trackers\cite{zhu2018distractor,li2018high,bertinetto2016fully,valmadre2017end,li2019siamvgg,he2018twofold} play an important role due to its satisfactory balance in speed and accuracy, especially in short-term realtime tracking.

Despite the performance gaining, applying CNN will also bring adversarial threats. Since the adversarial attack was first introduced by Szegedy et al.\cite{42503}, researchers have proposed several robust and efficient attack methods. These adversarial examples can not only unveil potential threats in the networks but also help network designers improve the robustness of networks. However, not much attention has been paid to the adversarial threats in visual tracking.

In this paper, we demonstrate the existence of adversarial threats in visual tracking by generating adversarial examples over Siamese trackers. There are two main challenges with generating adversarial examples for Siamese trackers: (1) Siamese trackers use the first frame of the video sequences to initialize the network. Therefore, we need to make the adversarial object not self-similar to achieve a successful attack, while attacking other Siamese-based visual tasks like face recognition only requires the adversarial object not to be similar to the original object. Suppressing $f(x_{adv}, x_{adv})$ is much more difficult than $f(x_{ori},x_{adv})$ when f is a Siamese network function. (2) A practical attack on Siamese trackers requires the created perturbation keeps consistent in the whole video sequence. Therefore, directly applying adversarial attacks on image sequences is unreasonable, considering about the uncertainties of viewing angels, background settings and so on.

In this paper, we propose the Siamese Tracker Attack (STA) algorithm, which can generate adversarial examples over Siamese trackers. 
By analyzing the weaknesses of Siamese trackers, we find that the excessive dependence on the similarity score and the side-effect of the cosine window penalty make Siamese trackers vulnerable to adversarial attacks on the similarity score. With a sufficiently low score, the cosine window penalty will mislead Siamese trackers, even make them completely drift off.
To make our adversarial examples practical, instead of directly applying STA on 2D video images, we design an end-to-end 3D texture generation pipeline to generate adversarial 3D objects. We also adopt the Expectation Over Transformation (EOT) framework \cite{DBLP:conf/icml/AthalyeEIK18} and model the uncertainties of viewing parameters within the optimization procedure. With 3D scenes and objects, the consistence of perturbation is naturally maintained so that we can evaluate the effectiveness of our algorithm from all viewing angles. In our experiments, we successfully reduce the tracking accuracy of three state-of-the-art Siamese trackers and even make them completely drift off. In addition, our experiments indicate that the asymmetrical network structure used in some Siamese trackers will make them less robust to our attack.

Our contributions can be summarized as follows:
\begin{itemize} 
   \item We analyze the weaknesses of Siamese trackers, revealing that the over-dependence on similarity score, the cosine window penalty, and the asymmetrical region proposal subnetwork will bring potential adversarial treats to Siamese trackers. To the best of our knowledge, this is the first work that focuses on the adversarial threats to visual trackers.
   \item On this basis, we propose STA algorithm, which generates robust 3D adversarial examples over Siamese trackers. To ensure the consistence of perturbations during the whole video sequences, we design an end-to-end 3D generation pipeline and build a 3D scene to produce video sequences for evaluation.
   \item We evaluate our attacks on three typical Siamese trackers. Results show that our adversarial examples successfully reduce the accuracy and robustness of Siamese trackers, and even make them drift off completely. We hope our analysis and attacks can provide some insights for Siamese tracker designers to avoid the vulnerabilities of Siamese trackers and improve their robustness.
\end{itemize}

\section{Related Work}
\subsection{Siamese Networks based Tracking}\label{Section:2.1}
Siamese trackers covert the tracking problem into a similarity learning problem. They use the initial appearance of the object as the exemplar image and try to locate the exemplar image in a larger search image from frame to frame. They apply an identical feature extraction to both the exemplar image and the search image, and then perform cross-correlation on the feature maps to measure their similarity in embedding space. The position of the maximum similarity score indicates the position of the target. Benefiting from avoiding online update for the deep network, Siamese trackers achieve success in balancing between accuracy and speed. It is notable that in VOT2018 Challenge\cite{kristan2018sixth}, eight of the top ten real-time short-term trackers are based on Siamese network. 

SiamFC\cite{bertinetto2016fully} is one of the pioneers\cite{held2016learning,tao2016siamese} that train a fully convolutional Siamese network to locate an exemplar image within a larger search image. Inherited from SiamFC, other Siamese trackers try to enhance the tracking accuracy and robustness by improving their network structure. SiamVGG\cite{li2019siamvgg} observes that AlexNet only provides limited feature extraction capabilities for the Siamese network. To improve the discrimination capability, it chooses the VGG-16\cite{simonyan2014very} network as its backbone CNN. SA-Siam\cite{he2018twofold} introduces a twofold Siamese network which consists of a semantic branch and an appearance branch. An attentional mechanism is used to make choices between two folds. SiamRPN\cite{li2018high} formulates the tracking process as a local one-shot detection framework and introduces a region proposal network (RPN) to generate accurate region proposals. DaSiamRPN\cite{zhu2018distractor} goes one step further by tackling the imbalanced distribution of training data. With effective training samples, DaSiamRPN is capable of distinguishing semantically similar targets so that it can be extended to perform long-term tracking.

\subsection{Robust Adversarial Attack}
Recently, adversarial attack has shown its power on attacking some computer vision tasks, including image classification\cite{moosavi2017universal}, object detection\cite{xie2017adversarial} and semantic segmentation\cite{arnab2018robustness,xie2017adversarial}. Follow-up works try to improve the robustness of adversarial examples. Kurakin et al.\cite{kurakin2016adversarial} generates printing digital adversarial examples and proves that a large portion of their adversarial examples fool the image classifier. Athalye et al.\cite{DBLP:conf/icml/AthalyeEIK18} transfers the attack models to 3D objects. They point out that the 3D rendering and the uncertainties of physical conditions should be modeled in the optimization process. Their EOT framework models transformations synthetically when generating adversarial perturbations and successfully attack an image classifier from the full perspectives with an adversarial 3D-printed object. Chen et al. \cite{chen2018shapeshifter} adapt EOT framework to adversarial attacks for object detection. They generate adversarial stop signs which successfully fool Faster R-CNN\cite{girshick2015fast}. Eykholt et al. \cite{eykholt2018robust} propose a general attack algorithm to train robust visual adversarial road signs under various environmental conditions such as different viewpoint. They also validate that their works can effectively attack both image classifier and object detector. Zhang et al.\cite{zhang2018camou} propose a framework called CAMOU to learn a camouflage pattern that can hide vehicles from being detected by state-of-the-art object detectors. They use a clone neural network to substitute the rendering engine and generate a camouflage that can minimize car detection scores by Mask R-CNN\cite{he2017mask} and YOLOv3-SPP\cite{redmon2018yolov3}.

However, not enough attention has been paid to attacking visual object trackers although visual object tracking is a fundamental problem in many computer topics such as visual analysis, automatic driving and pose estimation. 

\begin{figure}[t]
\begin{center}
  \includegraphics[width=1\linewidth]{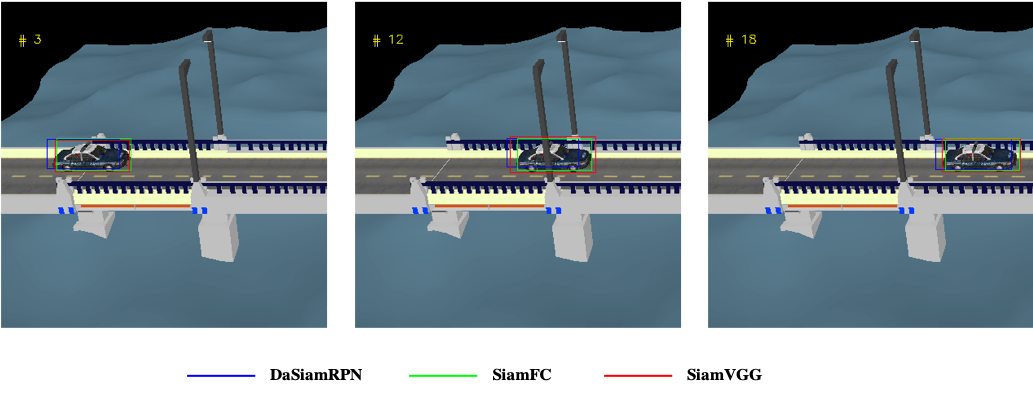}
\end{center}
   \caption{We test Siam-FC, Siam-VGG and Da-SiamRPN with a unperturbed car running on the bridge. Results show that occlusion caused by columns on the bridge has little effect on their tracking accuracy.}
\label{fig:3.1}
\end{figure}

\begin{table*}[h] \small
\caption{In this table we show the score maps before and after cosine window penalty. For the unperturbed car and rabbit, their scores (0.8/0.5 for the car/rabbit) are high enough so that they still keep highest after applying cosine window penalty. While for the adversarial objects, since their scores are suppressed severely (0.32/0.075 for the car/rabbit), they no longer maintain highest after applying cosine window penalty, leading to the failure of the tracker.}
\begin{center}
\setlength{\tabcolsep}{4mm}
\begin{tabular}{m{1cm}<{\centering}m{1.5cm}<{\centering}m{2cm}<{\centering}m{2cm}<{\centering}||
								   m{1.5cm}<{\centering}m{2cm}<{\centering}m{2cm}<{\centering}}
\specialrule{1pt}{3pt}{3pt}
Model & Image & Score Map (before Cos) & Score Map (after Cos) & Image & Score Map (before Cos) & Score Map (after Cos)\\ \specialrule{1pt}{1pt}{3pt}

Unperturbed Object & \includegraphics[width=1\linewidth]{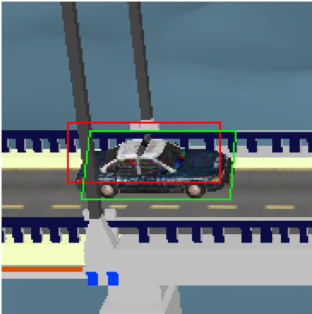} & \includegraphics[width=1\linewidth]{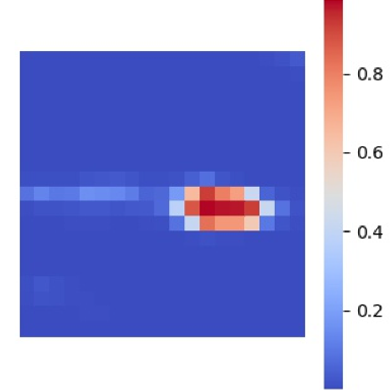} & \includegraphics[width=1\linewidth]{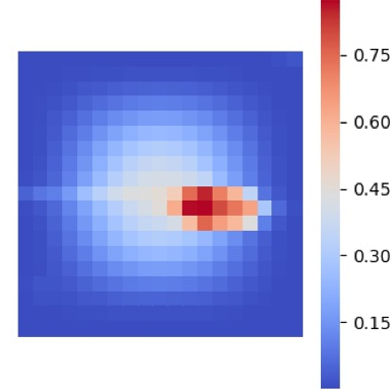} & \includegraphics[width=1\linewidth]{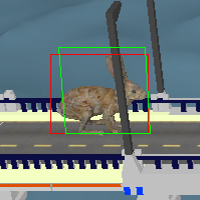} & \includegraphics[width=1\linewidth]{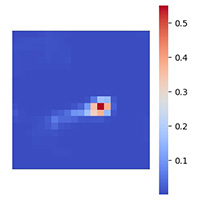} & \includegraphics[width=1\linewidth]{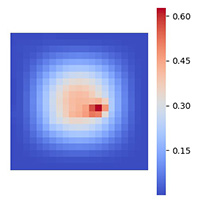} \\
\specialrule{1pt}{3pt}{3pt}
Adversarial Object & \includegraphics[width=1\linewidth]{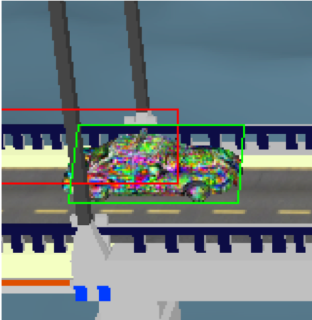}& \includegraphics[width=1\linewidth]{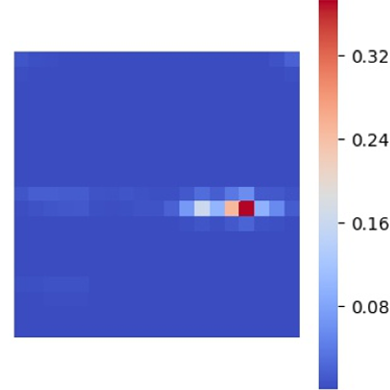} & \includegraphics[width=1\linewidth]{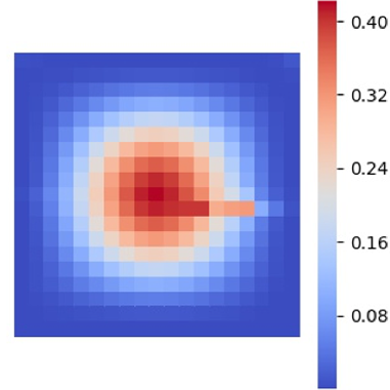} & \includegraphics[width=1\linewidth]{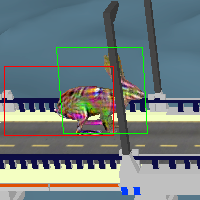}& \includegraphics[width=1\linewidth]{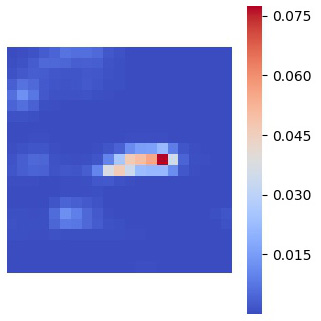} & \includegraphics[width=1\linewidth]{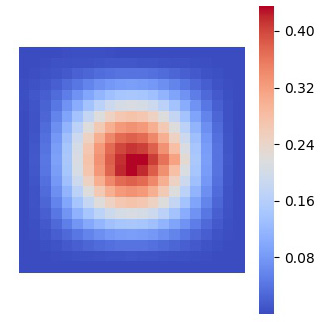} \\
\specialrule{1pt}{1pt}{1pt}
\end{tabular}
\end{center} 
\label{table:3.1}
\end{table*}

\section{Generating Adversarial Example}
In this section, we first analyze the weaknesses of Siamese trackers. On this basis, we present our Siamese Tracker Attack (STA) algorithm to generate robust adversarial 3D objects over Siamese trackers.

\subsection{Siamese Tracker Vulnerability Analysis}\label{Section:3.1}
Thanks to the powerful Siamese network and the large training dataset, Siamese trackers can locate the target robustly, even in some challenging situations. As Fig \ref{fig:3.1} shows, Siamese trackers successfully locate the target although it is occluded by a column on the bridge. 

However, there are some weaknesses in state-of-the-art Siamese trackers:

Siamese trackers highly rely on similarity score to predict the target's position. Other information like motion estimation and shape deformation is neglected. Such excessive dependence makes Siamese trackers vulnerable to adversarial attacks on the similarity score. For traditional Siamese trackers, although their complete symmetrical networks restrict us from suppressing the similarity score severely, we can still make them less robust to some challenging situations such as occlusion and deformation. While for RPN-based Siamese trackers, things get worse. Suffering from their asymmetrical networks, their score can be suppressed so severely that trackers can fail directly without any occlusion or deformation.

Furthermore, Siamese trackers introduce cosine window penalty to penalize the distractors that are far from the center\cite{zhu2018distractor}. Without cosine window penalty, Siamese trackers will be sensitive to any distractors or noises.  However, when we suppress the target's score through fabricating perturbed texture for the target, its side-effect appears. The gap between targets' score and background's score will become so small that cosine window will have a major impact on the final score and mislead the tracker. What's worse, when misleading happens, the target will be further away from the center of the cosine window, which will exaggerate the side-effect of cosine window penalty and even make the tracker drift off completely.

As we can see in Table \ref{table:3.1}, for the unperturbed car and rabbit, their scores are high enough so that they still keep highest after applying cosine window penalty. While for the adversarial objects, although their scores are still highest before applying cosine window penalty, it no longer maintains after applying the cosine window penalty, which eventually leads to the failure of Siamese trackers. 

Here, we will have a quantitative analysis on how much should we suppress the similarity score to achieve a successful attack. Supposing target's score is $s$, the score of the disturbing position is $s'$, target's distance from center is $d$, disturbing position's distance is $d'$, the penalty weight is $c$ and the total pixel number of score map is $M$, the final scores of target and disturbing position in Siamese trackers are:
\begin{equation}
\begin{split}
   s_{target} = (1-c) \cdot s + c \cdot [0.5 + 0.5cos(\frac{2 \pi d}{M-1})] \\
   s_{dispos} = (1-c) \cdot s' + c \cdot [0.5 + 0.5cos(\frac{2 \pi d'}{M-1})]
\end{split}
\end{equation}

To mislead the Siamese tracker, we need to make $s_{dispos} > s_{target}$. Substitute the expression in this inequality, and we can get when $\Delta s=s-s'<\frac{0.5c}{1-c} \cdot [cos(\frac{2 \pi d'}{M-1} - cos(\frac{2 \pi d} {M-1}))]$, the disturbing position will be misjudged as ground truth, leading to the failure of the Siamese trackers.

To sum up, a successful attack on Siamese trackers requires us to suppress the similarity score of the adversarial 3D object. A sufficiently low similarity score, plus the side-effect of cosine window penalty, will mislead the tracker and even make the tracker completely drift off.

\begin{figure*}
\begin{center}
   \includegraphics[width=0.75\linewidth]{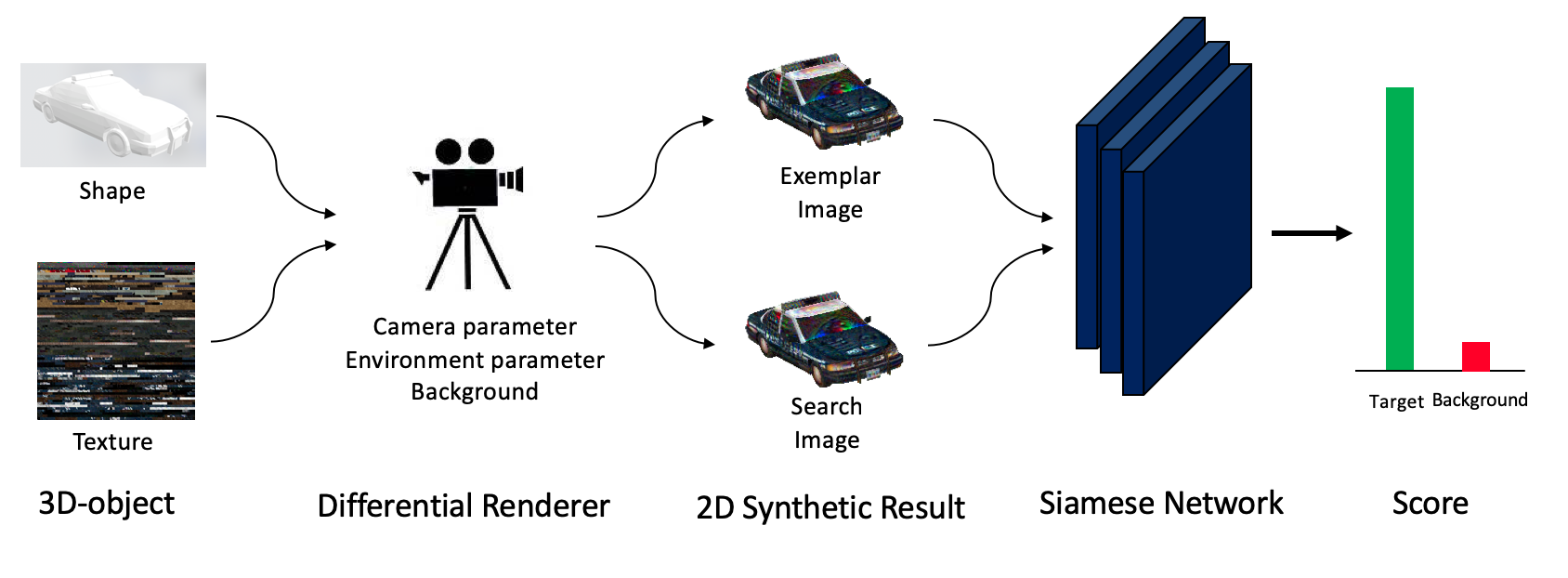}
\end{center}
   \caption{The forwarding pipeline of STA. We first use a differential renderer to render the 3D object into two images that both contain the target. Then we send these two images to the Siamese network and get their similarity score. Through backward propagation, STA generates the perturbed texture that minimizes the similarity score of the target.}
\label{fig:3.2}
\end{figure*}

\subsection{Siamese Tracker Attack}
Based on the analysis above, we formulate the adversarial generation problem and present STA algorithm to generate robust adversarial 3D object over Siamese trackers.

Given a specific Siamese tracker, we use $f(I_z,I_x)$ to denote its Siamese similarity scoring function. Since $f$ measures the similarity in image space, we first use a differentiable renderer to render the 3D object and get two images that both contain the target. To make the target not self-similar, we minimize the similarity score between them through generating a perturbed texture for the 3D object. Therefore, our optimizing model starts as:
\begin{equation}
\underset{\tilde{o_t}}{\arg\min} f(R(o_s,\tilde{o_t};p),R(o_s,\tilde{o_t};p))
\end{equation}
where $R(o;p)$ denotes a differentiable rendering process. Let $o = (o_s, o_t)$ be the 3D object. $o_s$ denotes its 3D-mesh and $o_t$ denotes its texture. With a collection of viewing parameters $p$, such as camera distance, viewing angle, lighting condition and solid background color, the 3D object is rendered into two images and then sent to similarity function.

It is notable that during the tracking process, trackers will scale the image to make sure that the size of the  bounding box, plus the context margin, is suitable for the Siamese network. For example, in SiamFC, given a tight bounding box with size $(w, h)$, the amount of context is set to be $p = (w + h)/4$. Siamese tracker will scale the image with scaling factor $s$ to make sure that $s(w + 2p) \times s(h + 2p) = A$, where A is the area of the exemplar images. 

It has been well demonstrated that such process will reduce the effectiveness of adversarial examples\cite{8294186}. To overcome it, we imitate the same cropping and scaling process to our rendering results before they are sent to the similarity function. We use $S$ to denote this process and define $T(o;p) = S(R(o;p))$. Then the optimization problem becomes:
\begin{equation}
\underset{\tilde{o_t}}{\arg\min} f(T(o_s,\tilde{o_t};p),T(o_s,\tilde{o_t};p))
\end{equation}

In practice, the 3D objects we generate can be tracked with different distances, angels, and scenes. To make our adversarial examples robust under most physical conditions, we adopt Expectation Over Transformation(EOT) framework proposed by Athalye et al.\cite{DBLP:conf/icml/AthalyeEIK18} and model such uncertainties within our optimization process.
With a distribution of viewing parameters $P$, we can minimize the score under expectation of distribution $P$. Therefore, our optimization problem can be described as:
\begin{equation}
\underset{\tilde{o_t}}{\arg\min} \mathbb{E}_{p \sim P} [f(T(o_s,\tilde{o_t};p),T(o_s,\tilde{o_t};p))]
\end{equation}

Furthermore, we introduce the L2 distance between the original texture map and the perturbed texture map to control the perceptibility of our adversarial examples:

\begin{equation}
   \begin{split}
   \underset{\tilde{o_t}}{\arg\min} \mathbb{E}_{p \sim P} [f(T(o_s,\tilde{o_t};p), T(o_s,\tilde{o_t};p)) \\
      + \lambda\cdot||\tilde{o_t} - o_t||_2]
   \end{split}
\end{equation}

To keep the perturbed texture map valid, we apply Projected Gradient Descent (PGD) to ensure the the value of the adversarial pixels is within [0, 1] and approximate the gradient of the expected value by sampling the viewing parameters independently in each forward-backward calculation. 

\subsection{Attack RPN-based Trackers} 
Currently, state-of-the-art Siamese trackers like SiamRPN\cite{li2018high} and DaSiamRPN\cite{zhu2018distractor} introduce Region Proposal Network (RPN) to improve tracking accuracy. Different from traditional Siamese trackers, these RPN-based trackers adopt a region proposal subnetwork for proposal generation instead of straightly performing correlation on feature maps. Unlike the similarity score in traditional Siamese trackers, RPN gives the foreground/background possibility for each anchor. 

To generate adversarial examples for RPN-based trackers, we attack their classification branch via misleading it to classify the target as background in all anchors. Based on our STA framework, we use cross-entropy $\mathcal{L}$ as the adversarial loss and the labels for all anchors $y^*$ are negative(background). Therefore, our optimization goal for RPN-based Siamese tracker becomes:
\begin{equation}
   \begin{split}
   \underset{\tilde{o_t}}{\arg\min} \mathbb{E}_{p \sim P} \{\mathcal{L}[f(T(o_s,\tilde{o_t};p), T(o_s,\tilde{o_t};p)),y^*] \\
      + \lambda\cdot||\tilde{o_t} - o_t||_2\}
   \end{split}
\end{equation}

\begin{figure}[t]
\begin{center}
  \includegraphics[width=0.8\linewidth]{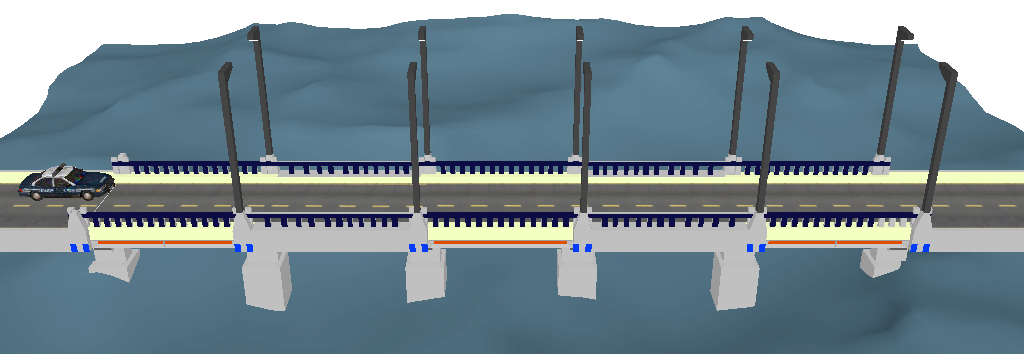}
\end{center}
   \caption{3D scene we build for evaluation. We set five columns on the bridge to evaluate the robustness of Siamese tracker. Occlusion happens when the car passes these columns.}
\label{fig:4.1}
\end{figure}

\begin{figure}[t]
\begin{center}
  \includegraphics[width=0.8\linewidth]{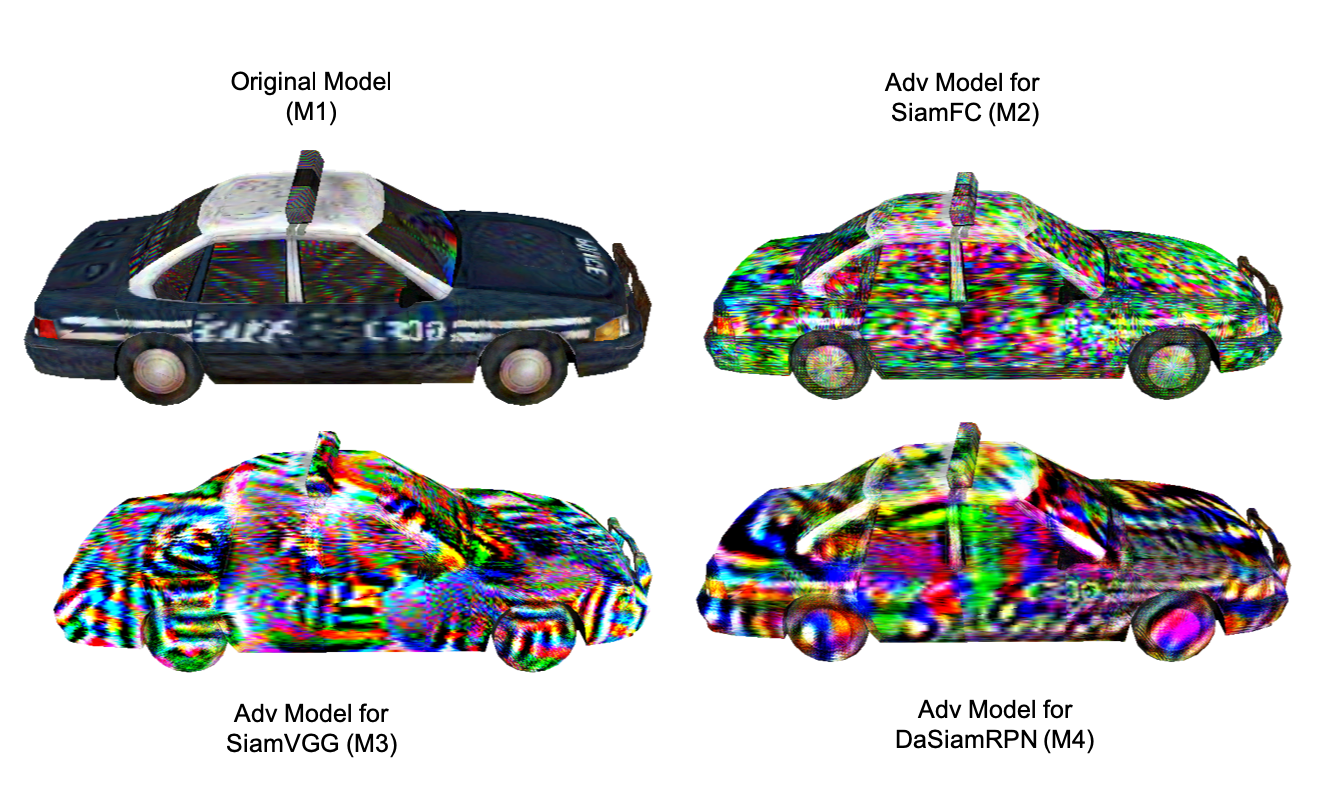}
\end{center}
   \caption{Adversarial cars generated by STA.}
\label{fig:4.2}
\end{figure}

\section{Experiments}

In this section, we first describe the Siamese trackers, attacking pipeline and evaluation metric we used in our experiment, and then we demonstrate how the adversarial examples we generate perform on three state-of-the-art Siamese trackers. After that, we discuss the side-effect of asymmetrical structure in RPN-based Siamese tracker, proving that this asymmetrical structure will reduce the robustness of the Siamese tracker. Finally, we investigate the transferability of the adversarial examples and try to generate one that can confuse all Siamese trackers we test.

\subsection{Experiment Setup}
\subsubsection{Software and Hardware Setup}
Since all three Siamese trackers we attack have pytorch\cite{paszke2017automatic} version, we use it as the deep learning framework for all experiments. Our experiments are performed on a Nvidia GTX1080Ti GPU. In terms of the differentiable renderer, we use neural 3D mesh renderer proposed by Hiroharu Kato et al. \cite{kato2018neural}.  

\subsubsection{The Siamese Trackers}
We choose SiamFC\cite{bertinetto2016fully}, SiamVGG\cite{li2019siamvgg} and DaSiamRPN\cite{zhu2018distractor} as our attack targets. SiamFC is the pioneer of Siamese trackers. Based on SiamFC, SiamVGG changes the backbone network into a modified VGG-16 network and achieve good results in VOT2018 challenge. These two trackers are traditional trackers which have a complete symmetrical structure. DaSiamRPN, an enhanced version of SiamRPN\cite{li2018high}, is the champion of the real-time short-term tracker in VOT2018. It represents the asymmetrical RPN-based trackers. We use the code released by its authors.

\subsubsection{Pipeline and Video Production}

Following the understanding of the problem we developed in previous sections, we build a generator whose pipeline is shown in Figure \ref{fig:3.2}. Furthermore, we build 3D scenes and produce 60-frame videos for both the original objects and the adversarial objects. The 3D scene we build is shown in Figure \ref{fig:4.1}. Occlusion happens when the objects pass by the columns on the bridge. Also, we mark down the ground truth in every frame. These videos are sent to the corresponding tracker for evaluation.

\subsubsection{Evaluation Metrics}

Three metrics are adopted to evaluate the performance of adversarial examples we generate. Two of them are quantitative. One is the drop of the highest similarity score since lowering target's similarity score is our direct attack goal. The other is the drop of Intersection Over Union (IOU). IoU between a predicted region $A$ and the ground truth region $B$ is defined as $IOU(A;B) = \frac{A\cap B} {A\cup B}$. It basically measures the accuracy of trackers and is widely used in popular video tracking benchmarks. Furthermore, we also adopt whether can adversarial examples make trackers drift off as our qualitative evaluation metric.

\begin{figure}[t]
\begin{center}
  \includegraphics[width=0.8\linewidth]{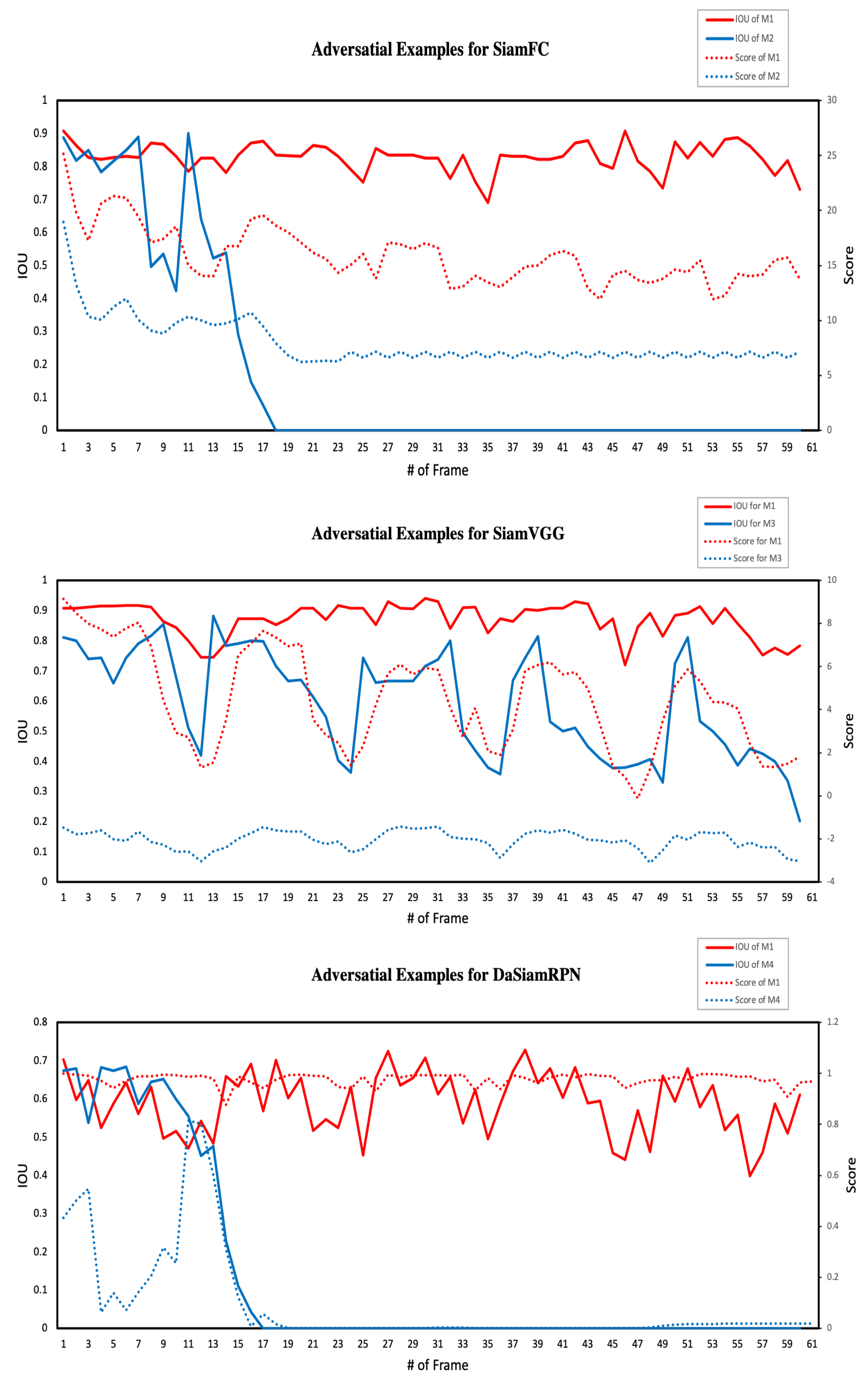}
\end{center}
   \caption{IOUs (solid lines) and Scores (dot lines) in three trackers. Red lines show the performance of original model and blue lines show the performance of adversarial models. For all three trackers, our attacks successfully decrease the similarity scores of the target. Also, we make these three trackers no longer robust to occlusion. SiamFC and SiamRPN completely drift off after the targets pass by the first column. SiamVGG meets sharp decrease when the target is occluded by the columns.}
\label{fig:4.3}
\end{figure}

\subsection{Attack Siamese Trackers}
In this section, we evaluate the performance of STA on three state-of-the-art Siamese trackers. Due to the limitation of pages, we only present the adversarial car here. More results, including attacks on other objects and the video clips we make can be found in the supplementary material. Figure \ref{fig:4.2} shows the adversarial examples generated by STA over three trackers. We observe that although models generated each time is different because of the randomness of viewing parameters, they seem to have a specific pattern according to the tracker they attack. Specifically, the texture for SiamFC looks like colorful spots on the car while textures for SiamVGG and SiamRPN look like two kinds of wave patterns on the car.

Figure \ref{fig:4.3} shows the comparison of IOUs (solid line) and similarity scores (dot line) of three trackers. As is shown in the figure, our attacks successfully decrease the similarity scores of all three trackers. More important, our adversarial examples make them no longer robust to occlusion.  When tracking the unperturbed car, all three trackers can track the car successfully with insignificant vibration in tracking accuracy. However, in terms of the adversarial cars, SiamFC and DaSiamRPN drift off directly when the car is occluded by the first column on the bridge. Although SiamVGG shows better robustness that it does not drift off in the whole video, its accuracy still meets sharp decrease every time the adversarial car passes the columns on the bridge. Five sharp decreases of IOU can be observed in the result of SiamVGG, which correspond to five columns on the bridge.

\begin{figure}[t]
\begin{center}
  \includegraphics[width=0.8\linewidth]{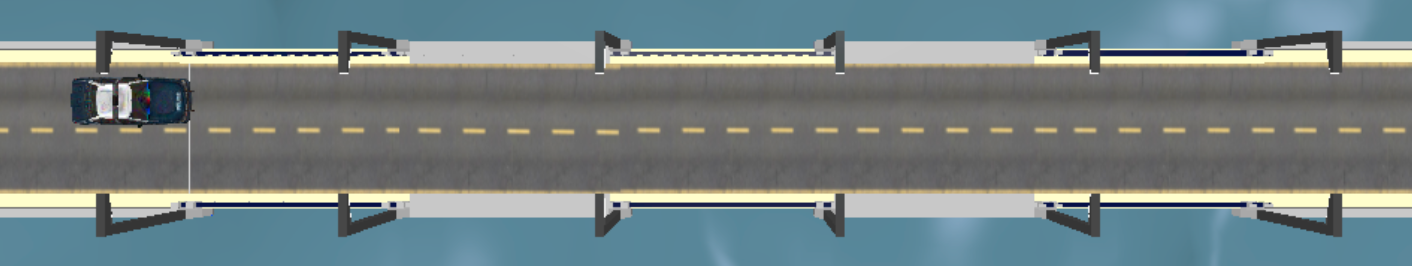}
\end{center}
   \caption{From this viewing angle, the car will no be occluded by the columns on the bridge in the whole video sequences.}
\label{fig:new1}
\end{figure}

\begin{table*}[t] \small
\caption{Results for SiamFC and DaSiamRPN. For SiamFC, although we successfully suppress its similarity score, we fail to make it drift off since its similarity score is still high enough to overcome the side-effect of cosine window. However, our adversarial object successfully fools DaSiamRPN without occlusion. Its score is suppressed to 0.1 so that it is no longer recognized after cosine window penalty. (Range of score in DaSiamRPN is 0 to 1. In the same scene, the score of the original model is 0.99)}
\begin{center}
\setlength{\tabcolsep}{4mm}
\begin{tabular}{m{0.2cm}<{\centering}m{1.5cm}<{\centering}m{2cm}<{\centering}m{2cm}<{\centering} |
m{0.2cm}<{\centering}m{1.5cm}<{\centering}m{2cm}<{\centering}m{2cm}<{\centering}}

\specialrule{1pt}{0pt}{0pt}
\multicolumn{4}{c}{SiamFC} & \multicolumn{4}{c}{DaSiamRPN}
\\ \specialrule{1pt}{1pt}{3pt}

Frame No. & Image & Scoremap (before Cos) & Scoremap (after Cos)
&Frame No. & Image & Scoremap (before Cos) & Scoremap (after Cos)\\ \specialrule{1pt}{1pt}{3pt}

5 & \includegraphics[width=1\linewidth]{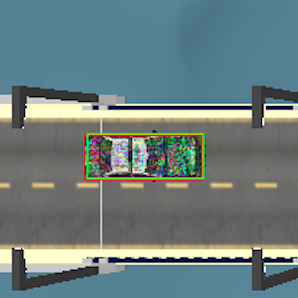} & \includegraphics[width=1\linewidth]{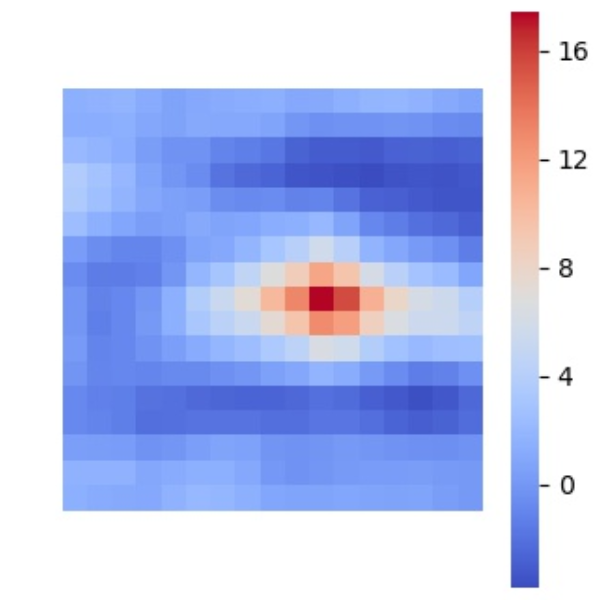} & \includegraphics[width=1\linewidth]{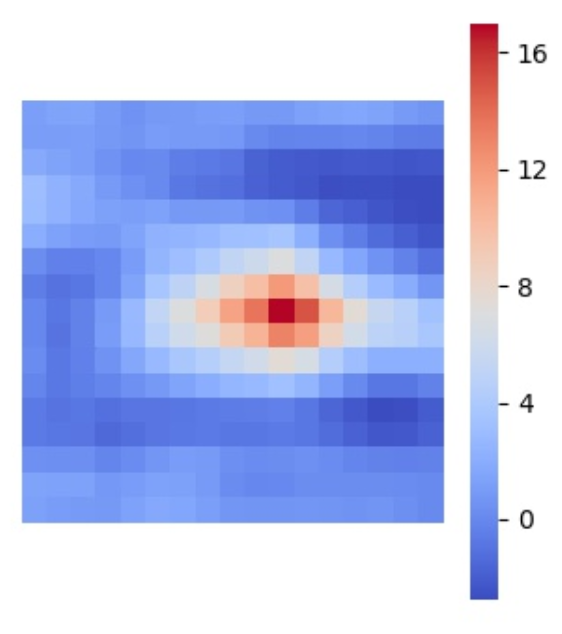} &
5 & \includegraphics[width=1\linewidth]{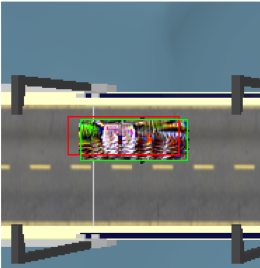} & \includegraphics[width=1\linewidth]{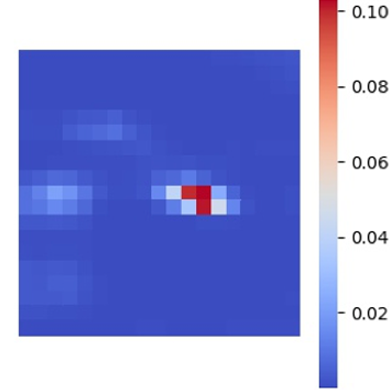} & \includegraphics[width=1\linewidth]{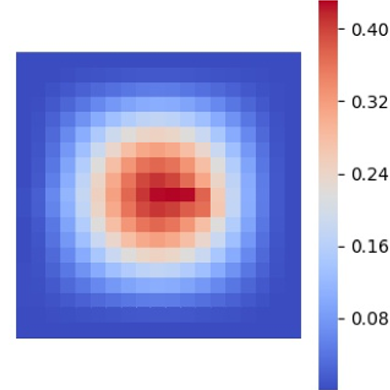}  \\\specialrule{1pt}{3pt}{3pt}

7 & \includegraphics[width=1\linewidth]{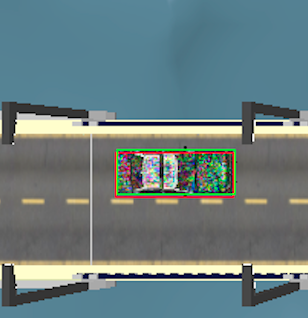}& \includegraphics[width=1\linewidth]{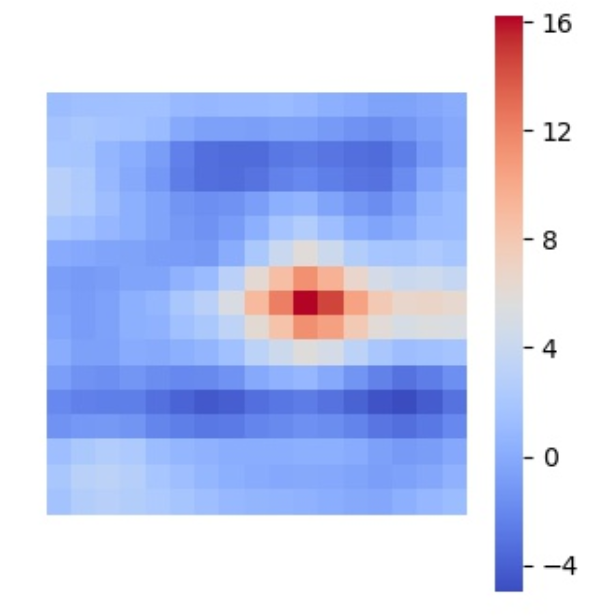} & \includegraphics[width=1\linewidth]{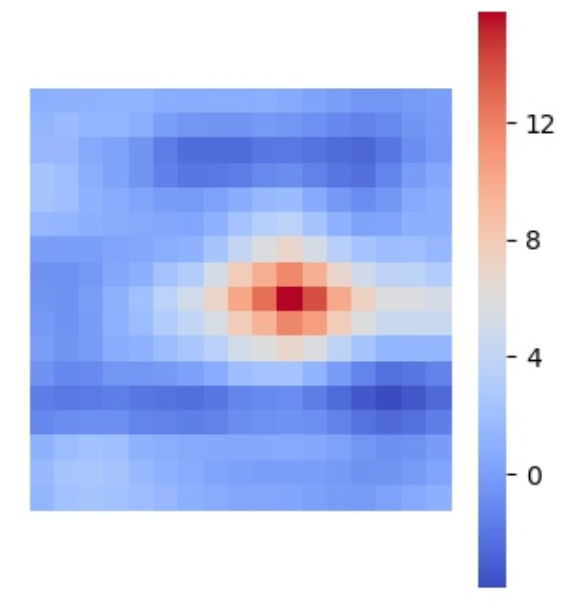} &
7 & \includegraphics[width=1\linewidth]{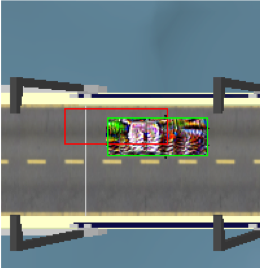}& \includegraphics[width=1\linewidth]{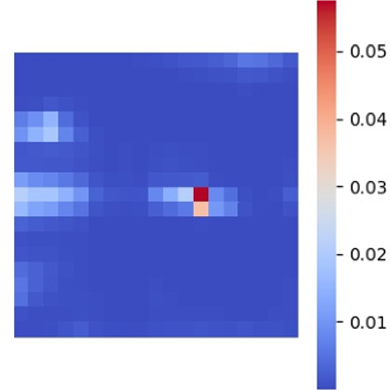} & \includegraphics[width=1\linewidth]{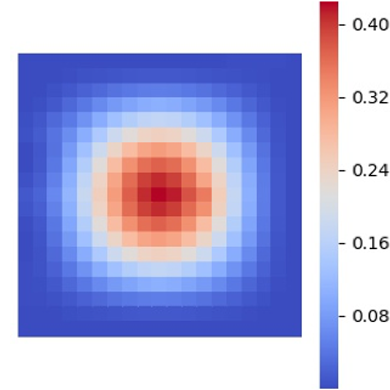}  \\ \specialrule{1pt}{3pt}{3pt}

9 & \includegraphics[width=1\linewidth]{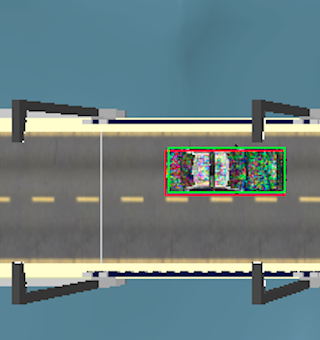}& \includegraphics[width=1\linewidth]{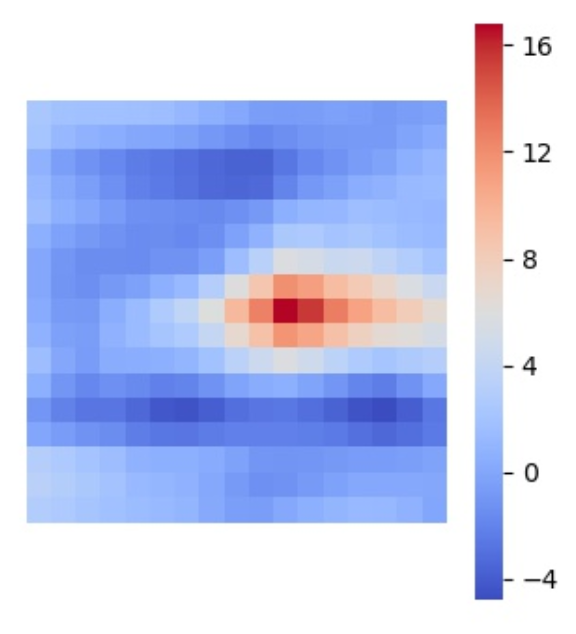} & \includegraphics[width=1\linewidth]{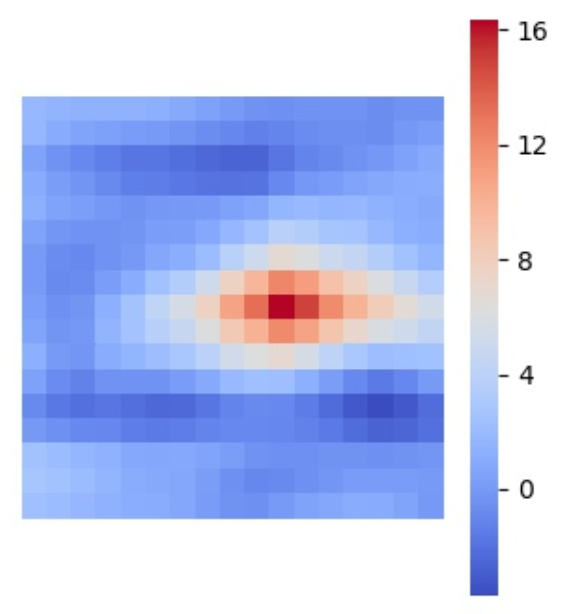} &
9 & \includegraphics[width=1\linewidth]{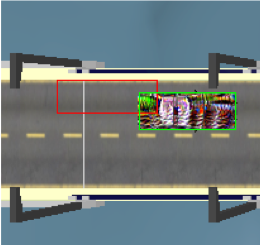}& \includegraphics[width=1\linewidth]{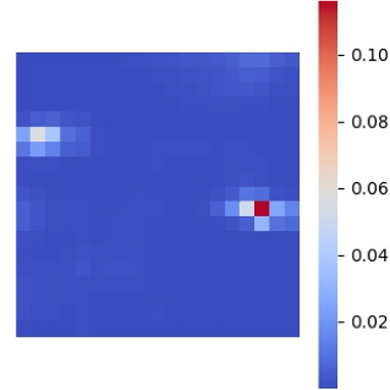} & \includegraphics[width=1\linewidth]{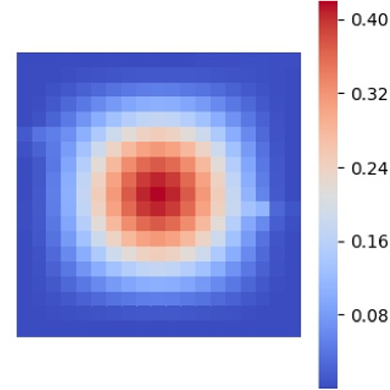}\\ \specialrule{1pt}{3pt}{0pt}

\end{tabular}
\end{center}

\label{table:4.2}
\end{table*}

\begin{table*}[t] \small
\caption{mIOU(\%) results of transferability experiments. Columns 2-4 show how the adversarial examples generated from one tracker perform on other trackers. Columns 5-8 show the effectiveness of combining attack.}  
\begin{center}
\begin{tabular}{|c|c|c|c|c|c|c|c|c|c|}
\hline
\multirow{2}*{Model} & \multicolumn{9}{c|}{Adversarial Examples Generated from} \\ 
\cline{2-10}
     & \bf{SiamFC(r1)} & \bf{SiamVGG(r2)} & \bf{DaSiamRPN(r3)} & \bf{r1+r2} & \bf{r1+r3} & \bf{r2+r3} & \bf{r1+r2+r3} & \bf{none}& \bf{random}\\
\hline 
\bf{SiamFC} & \bf{17.44} & 83.83 & 86.24 & \bf{10.07} & \bf{31.66} &  87.68 & \bf{10.3} & 82.78 & 83.36\\
\cdashline{1-10}[0.8pt/2pt]
\bf{SiamVGG}  & 81.67 & \bf{59.42} & 87.9 & \bf{66.73} & 79.76 & \bf{75.75} & \bf{76.69} & 86.99 & 86.27\\
\cdashline{1-10}[0.8pt/2pt]
\bf{DaSiamRPN}  & 58.06 & 58.14 & \bf{13.7} & 58.66 & \bf{14.91} & \bf{14.76} & \bf{14.71} & 58.98 & 56.74\\
\hline

\end{tabular}
\end{center}
 
\label{tab:table2}
\end{table*}

\subsection{Asymmetrical Network Breeds Danger}
According to the analysis before, to achieve a successful attack, the similarity score has to be suppressed sufficiently. Also, it is suggested that the asymmetrical structure of RPN-based Siamese trackers breeds the possibility to suppress the score severely. In this section, to prove the vulnerabilities brought by this asymmetrical structure, we render the 3D scene from another viewing angle. As is shown in figure \ref{fig:new1}, the target car is not occluded in the whole video sequences from this angle. Our goals is to test whether STA can make SiamFC and DaSiamRPN fail without occlusion, under the same hyper parameter settings.

Experiment results validate the vulnerability of this asymmetrical structure. When we attack SiamFC, although we can suppress the similarity score to some extent, we fail to make it drift off in the whole video sequences. However, in terms of DaSiamRPN, STA succeeds in making it drift off without any occlusion.

Detail score maps in Table \ref{table:4.2} can explain this result. As we can see, the score of the adversarial target in SiamFC remains highest after cosine window penalty because its symmetrical structure prevents its similarity score from severely decreasing. While in terms of DaSiamRPN, the score of our adversarial example is suppressed to be so low that it drowns in the cosine window, leading to the failure of DaSiamRPN.

\subsection{Transferability Experiment}
In this section, we investigate the transferability of our adversarial examples. To this end, we feed the adversarial examples we generate for each tracker to the other trackers. For comparison, we also use these trackers to track a car with totally random noise. Furthermore, we try to conduct combining attack by using STA on three trackers iteratively with a decreasing learning rate.

Table \ref{tab:table2} indicates the lack of transferability of adversarial examples generated by STA. The adversarial object we generate for one tracker fails to show noticeable effects on the other two trackers. This result is not surprising since previous researchers has argued that high-confidence physical attacks fail to transfer both in image recognition and object detection\cite{DBLP:journals/corr/LiuCLS16,chen2018shapeshifter}. 
 
In terms of combining attack, it works on these three trackers. As results in columns 5-8 show, through applying STA on three trackers iteratively with a decreasing learning rate. We successfully fool three trackers with a single adversarial object.

\section{Discussion and Future Work}
\noindent \textbf{Perceptibility.}
In the course of our experiments, to generate a robust adversarial 3D object, we use a very small $\lambda$ so that the perturbation of texture is quite perceptible, as we can see in Figure \ref{fig:4.2}. Although adversarial attacks over other tasks require perturbation to be unnoticeable, we think it acceptable in this case since the shape of the object is kept and what STA does can be considered as designing camouflages for the objects. 

\medskip
\noindent \textbf{Limits on Experiments.}
We have not evaluated our adversarial examples in the physical world due to the financial limitation. Also, state-of-the-art differentiable renderer still cannot tackle complicated background and produce high-resolution images, which limits the robustness of our adversarial examples. 

Furthermore, we do not attack popular video datasets such as VOT and OTB because these datasets only provide 2D image sequences. As is mentioned before, a practical attack on video trackers can not be conducted with 2D images. A 3D video tracking dataset or the 3D reconstruction technology may be the possible solution.

\medskip
\noindent \textbf{Suggestions on Designing Siamese Trackers.}
First and the most important, attentions should be paid to the side-effect of cosine window penalty. In existing Siamese trackers, cosine window is located at the predicted position in the last frame so that the target will always be punished by cosine window. Simply removing the cosine window penalty will make trackers not robust to distractors or sharp noises. Therefore, it is suggested to improve the location of the cosine window using motion estimation. With more accurate center, the side-effect of cosine window can be eased. Also, a completely symmetrical network structure is suggested for Siamese trackers because of its robustness to adversarial attacks.

\section{Conclusion}
In this paper, we demonstrate the potential adversarial threats on Siamese trackers and propose STA algorithm to generate adversarial 3D objects over three state-of-the-art Siamese trackers. Experiment results show that our adversarial examples successfully decrease their tracking accuracy and even make them drift off completely. Furthermore, the asymmetrical structure of RPN-based Siamese trackers makes them easier to be attacked. 

To conclude, the over-dependence on similarity score, the side-effect of existing cosine window penalty, and the asymmetrical region proposal subnetwork will make Siamese trackers less robust to adversarial attacks, which is suggested to be considered by Siamese tracker designers.


\begin{thebibliography}{}

\bibitem[\protect\citeauthoryear{{Akhtar} and {Mian}}{2018}]{8294186}
{Akhtar}, N., and {Mian}, A.
\newblock 2018.
\newblock Threat of adversarial attacks on deep learning in computer vision: A
  survey.
\newblock {\em IEEE Access} 6:14410--14430.

\bibitem[\protect\citeauthoryear{Arnab, Miksik, and
  Torr}{2018}]{arnab2018robustness}
Arnab, A.; Miksik, O.; and Torr, P.~H.
\newblock 2018.
\newblock On the robustness of semantic segmentation models to adversarial
  attacks.
\newblock In {\em Proceedings of the IEEE Conference on Computer Vision and
  Pattern Recognition},  888--897.

\bibitem[\protect\citeauthoryear{Athalye \bgroup et al\mbox.\egroup
  }{2018}]{DBLP:conf/icml/AthalyeEIK18}
Athalye, A.; Engstrom, L.; Ilyas, A.; and Kwok, K.
\newblock 2018.
\newblock Synthesizing robust adversarial examples.
\newblock In {\em Proceedings of the 35th International Conference on Machine
  Learning, {ICML} 2018, Stockholmsm{\"{a}}ssan, Stockholm, Sweden, July 10-15,
  2018},  284--293.

\bibitem[\protect\citeauthoryear{Bertinetto \bgroup et al\mbox.\egroup
  }{2016}]{bertinetto2016fully}
Bertinetto, L.; Valmadre, J.; Henriques, J.~F.; Vedaldi, A.; and Torr, P.~H.
\newblock 2016.
\newblock Fully-convolutional siamese networks for object tracking.
\newblock In {\em European conference on computer vision},  850--865.
\newblock Springer.

\bibitem[\protect\citeauthoryear{Chen \bgroup et al\mbox.\egroup
  }{2018}]{chen2018shapeshifter}
Chen, S.-T.; Cornelius, C.; Martin, J.; and Chau, D. H.~P.
\newblock 2018.
\newblock Shapeshifter: Robust physical adversarial attack on faster r-cnn
  object detector.
\newblock In {\em Joint European Conference on Machine Learning and Knowledge
  Discovery in Databases},  52--68.
\newblock Springer.

\bibitem[\protect\citeauthoryear{Eykholt \bgroup et al\mbox.\egroup
  }{2018}]{eykholt2018robust}
Eykholt, K.; Evtimov, I.; Fernandes, E.; Li, B.; Rahmati, A.; Xiao, C.;
  Prakash, A.; Kohno, T.; and Song, D.
\newblock 2018.
\newblock Robust physical-world attacks on deep learning visual classification.
\newblock In {\em Proceedings of the IEEE Conference on Computer Vision and
  Pattern Recognition},  1625--1634.

\bibitem[\protect\citeauthoryear{Girshick}{2015}]{girshick2015fast}
Girshick, R.
\newblock 2015.
\newblock Fast r-cnn.
\newblock In {\em Proceedings of the IEEE international conference on computer
  vision},  1440--1448.

\bibitem[\protect\citeauthoryear{He \bgroup et al\mbox.\egroup
  }{2017}]{he2017mask}
He, K.; Gkioxari, G.; Doll{\'a}r, P.; and Girshick, R.
\newblock 2017.
\newblock Mask r-cnn.
\newblock In {\em Proceedings of the IEEE international conference on computer
  vision},  2961--2969.

\bibitem[\protect\citeauthoryear{He \bgroup et al\mbox.\egroup
  }{2018}]{he2018twofold}
He, A.; Luo, C.; Tian, X.; and Zeng, W.
\newblock 2018.
\newblock A twofold siamese network for real-time object tracking.
\newblock In {\em Proceedings of the IEEE Conference on Computer Vision and
  Pattern Recognition},  4834--4843.

\bibitem[\protect\citeauthoryear{Held, Thrun, and
  Savarese}{2016}]{held2016learning}
Held, D.; Thrun, S.; and Savarese, S.
\newblock 2016.
\newblock Learning to track at 100 fps with deep regression networks.
\newblock In {\em European Conference on Computer Vision},  749--765.
\newblock Springer.

\bibitem[\protect\citeauthoryear{Kato, Ushiku, and
  Harada}{2018}]{kato2018neural}
Kato, H.; Ushiku, Y.; and Harada, T.
\newblock 2018.
\newblock Neural 3d mesh renderer.
\newblock In {\em Proceedings of the IEEE Conference on Computer Vision and
  Pattern Recognition},  3907--3916.

\bibitem[\protect\citeauthoryear{Kristan \bgroup et al\mbox.\egroup
  }{2018}]{kristan2018sixth}
Kristan, M.; Leonardis, A.; Matas, J.; Felsberg, M.; Pflugfelder, R.; Zajc,
  L.~{\v{C}}.; Voj{\'\i}r, T.; Bhat, G.; Luke{\v{z}}i{\v{c}}, A.; Eldesokey,
  A.; et~al.
\newblock 2018.
\newblock The sixth visual object tracking vot2018 challenge results.
\newblock In {\em European Conference on Computer Vision},  3--53.
\newblock Springer.

\bibitem[\protect\citeauthoryear{Krizhevsky, Sutskever, and
  Hinton}{2012}]{krizhevsky2012imagenet}
Krizhevsky, A.; Sutskever, I.; and Hinton, G.~E.
\newblock 2012.
\newblock Imagenet classification with deep convolutional neural networks.
\newblock In {\em Advances in neural information processing systems},
  1097--1105.

\bibitem[\protect\citeauthoryear{Kurakin, Goodfellow, and
  Bengio}{2016}]{kurakin2016adversarial}
Kurakin, A.; Goodfellow, I.; and Bengio, S.
\newblock 2016.
\newblock Adversarial examples in the physical world.
\newblock {\em arXiv preprint arXiv:1607.02533}.

\bibitem[\protect\citeauthoryear{Li and Zhang}{2019}]{li2019siamvgg}
Li, Y., and Zhang, X.
\newblock 2019.
\newblock Siamvgg: Visual tracking using deeper siamese networks.
\newblock {\em arXiv preprint arXiv:1902.02804}.

\bibitem[\protect\citeauthoryear{Li \bgroup et al\mbox.\egroup
  }{2018}]{li2018high}
Li, B.; Yan, J.; Wu, W.; Zhu, Z.; and Hu, X.
\newblock 2018.
\newblock High performance visual tracking with siamese region proposal
  network.
\newblock In {\em Proceedings of the IEEE Conference on Computer Vision and
  Pattern Recognition},  8971--8980.

\bibitem[\protect\citeauthoryear{Liu \bgroup et al\mbox.\egroup
  }{2016}]{DBLP:journals/corr/LiuCLS16}
Liu, Y.; Chen, X.; Liu, C.; and Song, D.
\newblock 2016.
\newblock Delving into transferable adversarial examples and black-box attacks.
\newblock {\em CoRR} abs/1611.02770.

\bibitem[\protect\citeauthoryear{Moosavi-Dezfooli \bgroup et al\mbox.\egroup
  }{2017}]{moosavi2017universal}
Moosavi-Dezfooli, S.-M.; Fawzi, A.; Fawzi, O.; and Frossard, P.
\newblock 2017.
\newblock Universal adversarial perturbations.
\newblock In {\em Proceedings of the IEEE Conference on Computer Vision and
  Pattern Recognition},  1765--1773.

\bibitem[\protect\citeauthoryear{Paszke \bgroup et al\mbox.\egroup
  }{2017}]{paszke2017automatic}
Paszke, A.; Gross, S.; Chintala, S.; Chanan, G.; Yang, E.; DeVito, Z.; Lin, Z.;
  Desmaison, A.; Antiga, L.; and Lerer, A.
\newblock 2017.
\newblock Automatic differentiation in pytorch.

\bibitem[\protect\citeauthoryear{Redmon and Farhadi}{2018}]{redmon2018yolov3}
Redmon, J., and Farhadi, A.
\newblock 2018.
\newblock Yolov3: An incremental improvement.
\newblock {\em arXiv preprint arXiv:1804.02767}.

\bibitem[\protect\citeauthoryear{Simonyan and
  Zisserman}{2014}]{simonyan2014very}
Simonyan, K., and Zisserman, A.
\newblock 2014.
\newblock Very deep convolutional networks for large-scale image recognition.
\newblock {\em arXiv preprint arXiv:1409.1556}.

\bibitem[\protect\citeauthoryear{Szegedy \bgroup et al\mbox.\egroup
  }{2014}]{42503}
Szegedy, C.; Zaremba, W.; Sutskever, I.; Bruna, J.; Erhan, D.; Goodfellow, I.;
  and Fergus, R.
\newblock 2014.
\newblock Intriguing properties of neural networks.
\newblock In {\em International Conference on Learning Representations}.

\bibitem[\protect\citeauthoryear{Tao, Gavves, and
  Smeulders}{2016}]{tao2016siamese}
Tao, R.; Gavves, E.; and Smeulders, A.~W.
\newblock 2016.
\newblock Siamese instance search for tracking.
\newblock In {\em Proceedings of the IEEE conference on computer vision and
  pattern recognition},  1420--1429.

\bibitem[\protect\citeauthoryear{Valmadre \bgroup et al\mbox.\egroup
  }{2017}]{valmadre2017end}
Valmadre, J.; Bertinetto, L.; Henriques, J.; Vedaldi, A.; and Torr, P.~H.
\newblock 2017.
\newblock End-to-end representation learning for correlation filter based
  tracking.
\newblock In {\em Proceedings of the IEEE Conference on Computer Vision and
  Pattern Recognition},  2805--2813.

\bibitem[\protect\citeauthoryear{Xie \bgroup et al\mbox.\egroup
  }{2017}]{xie2017adversarial}
Xie, C.; Wang, J.; Zhang, Z.; Zhou, Y.; Xie, L.; and Yuille, A.
\newblock 2017.
\newblock Adversarial examples for semantic segmentation and object detection.
\newblock In {\em Proceedings of the IEEE International Conference on Computer
  Vision},  1369--1378.

\bibitem[\protect\citeauthoryear{Zhang \bgroup et al\mbox.\egroup
  }{2018}]{zhang2018camou}
Zhang, Y.; Foroosh, H.; David, P.; and Gong, B.
\newblock 2018.
\newblock Camou: Learning physical vehicle camouflages to adversarially attack
  detectors in the wild.

\bibitem[\protect\citeauthoryear{Zhu \bgroup et al\mbox.\egroup
  }{2018}]{zhu2018distractor}
Zhu, Z.; Wang, Q.; Li, B.; Wu, W.; Yan, J.; and Hu, W.
\newblock 2018.
\newblock Distractor-aware siamese networks for visual object tracking.
\newblock In {\em Proceedings of the European Conference on Computer Vision
  (ECCV)},  101--117.

\end{thebibliography}
\end{document}